%% file: root.tex
\documentclass[letterpaper, 10 pt, conference]{ieeeconf}  

\IEEEoverridecommandlockouts                              

\usepackage{amsmath} 
\usepackage{bbm}
\usepackage{dsfont}
\usepackage{algorithm}
\usepackage{algorithmic}

\newcommand{\ours}{BehAV} 
\long\def\invis#1{}

\input{macros.tex}
\title{\LARGE \bf  \ours: Behavioral Rule Guided Autonomy Using VLMs for Robot Navigation in Outdoor Scenes
}

\author{Kasun Weerakoon$^1$, Mohamed Elnoor$^1$, Gershom Seneviratne$^1$, Vignesh Rajagopal$^3$, Senthil Hariharan Arul$^1$, \\ Jing Liang$^2$, Mohamed Khalid M Jaffar$^4$, and Dinesh Manocha$^{1,2}$ 
\thanks{This work was supported in part by ARO Grant W911NF2310352 and Army Cooperative Agreement W911NF2120076. We acknowledge the support of the Maryland Robotics Center.}
\thanks{$^{1}$Authors are with Dept. of Electrical and Computer Engineering, University of Maryland, College Park, MD, USA. {\tt\footnotesize kasunw@umd.edu, melnoor@umd.edu, gershom@umd.edu, sarul1@umd.edu}}
\thanks{$^{2}$ Authors are with Dept. of Computer Science, University of Maryland, College Park, MD, USA. {\tt\footnotesize jingl@umd.edu, dm@cs.umd.edu}} 
\thanks{$^{3}$ Author is with A. James Clark School of Engineering, University of Maryland, College Park, MD, USA. {\tt\footnotesize vigneshr@umd.edu}} 
\thanks{$^{4}$ Author is with Dept. of Aerospace Engineering, University of Maryland, College Park, MD, USA. {\tt\footnotesize khalid26@umd.edu}}\\
}

\begin{document}

\maketitle
\thispagestyle{empty}
\pagestyle{empty}

\begin{abstract}

We present \ours, a novel approach for autonomous robot navigation in outdoor scenes guided by human instructions and leveraging Vision Language Models (VLMs). Our method interprets human commands using a Large Language Model (LLM), and categorizes the instructions into navigation and behavioral guidelines. Navigation guidelines consist of directional commands (e.g., \textit{``move forward until"}) and associated landmarks (e.g., \textit{``the building with blue windows"}), while behavioral guidelines encompass regulatory actions (e.g., \textit{``stay on"}) and their corresponding objects (e.g., \textit{``pavements"}). We use VLMs for their zero-shot scene understanding capabilities to estimate landmark locations from RGB images for robot navigation. Further, we introduce a novel scene representation that utilizes VLMs to ground behavioral rules into a behavioral cost map. This cost map encodes the presence of behavioral objects within the scene and assigns costs based on their regulatory actions. The behavioral cost map is integrated with a LiDAR-based occupancy map for navigation. To navigate outdoor scenes while adhering to the instructed behaviors, we present an unconstrained Model Predictive Control (MPC)-based planner that prioritizes both reaching landmarks and following behavioral guidelines. We evaluate the performance of \ours{} on a quadruped robot across diverse real-world scenarios, demonstrating a {22.49}\% improvement in alignment with human-teleoperated actions, as measured by Fréchet distance, and achieving a {40}\% higher navigation success rate compared to state-of-the-art methods. Code and video are available at \url{http://gamma.umd.edu/behav/}.


\end{abstract}

\input{1_Introduction}
\input{2_Related_Work}

\input{3_Background}
\input{4_Our_Method}
\input{5_Results}
\input{6_Conclusions}

\bibliographystyle{IEEEtran}
\bibliography{References}

\end{document}

%% file: macros.tex
\usepackage{xcolor}
\usepackage{amssymb}
\usepackage{graphicx}
\usepackage{mathtools}
\usepackage{amsmath}
\usepackage{gensymb}
\usepackage{float}
\usepackage{multirow}
\usepackage{makecell}
\usepackage{mathtools}
\usepackage{hyperref}
\hypersetup{
    colorlinks=true,
    linkcolor=blue,
    filecolor=magenta,      
    urlcolor=blue,
    pdftitle={Overleaf Example},
    pdfpagemode=FullScreen,
    }

\usepackage{tabularx}
    \newcolumntype{L}{>{\raggedright\arraybackslash}X}
\usepackage[utf8]{inputenc}
\usepackage[english]{babel}

\usepackage{amsthm}

\usepackage{subcaption}

\newcommand{\no}{\noindent}



%% file: 1_Introduction.tex
\section{Introduction} \label{sec:intro}

Mobile robot navigation has gained significant attention in the last few decades due to its usability in numerous outdoor applications including delivery \cite{lee2021delivery}, inspection \cite{wang2020inspection}, surveillance \cite{sathyamoorthy2021surveillance}, search and rescue \cite{silver2010applied}, etc. Such applications often require the robots to follow scene-specific behaviors to achieve the navigation objectives. For example, in urban environments, a robot may need to yield to cyclists, remain on paved pathways, and comply with traffic signals in construction zones. Consequently, the robot must be capable of perceiving key behavioral objects within a scene (e.g., \textit{``Stop Sign", ``Paved Region''}) and understanding the corresponding actions required (e.g., \textit{``stop", ``stay on"}) to navigate effectively. Moreover, in highly dynamic environments, real-time scene perception becomes crucial for the robot to avoid potential collisions and prevent behaviorally inappropriate actions.

\begin{figure}[t]
      \centering
      \includegraphics[width=\columnwidth]{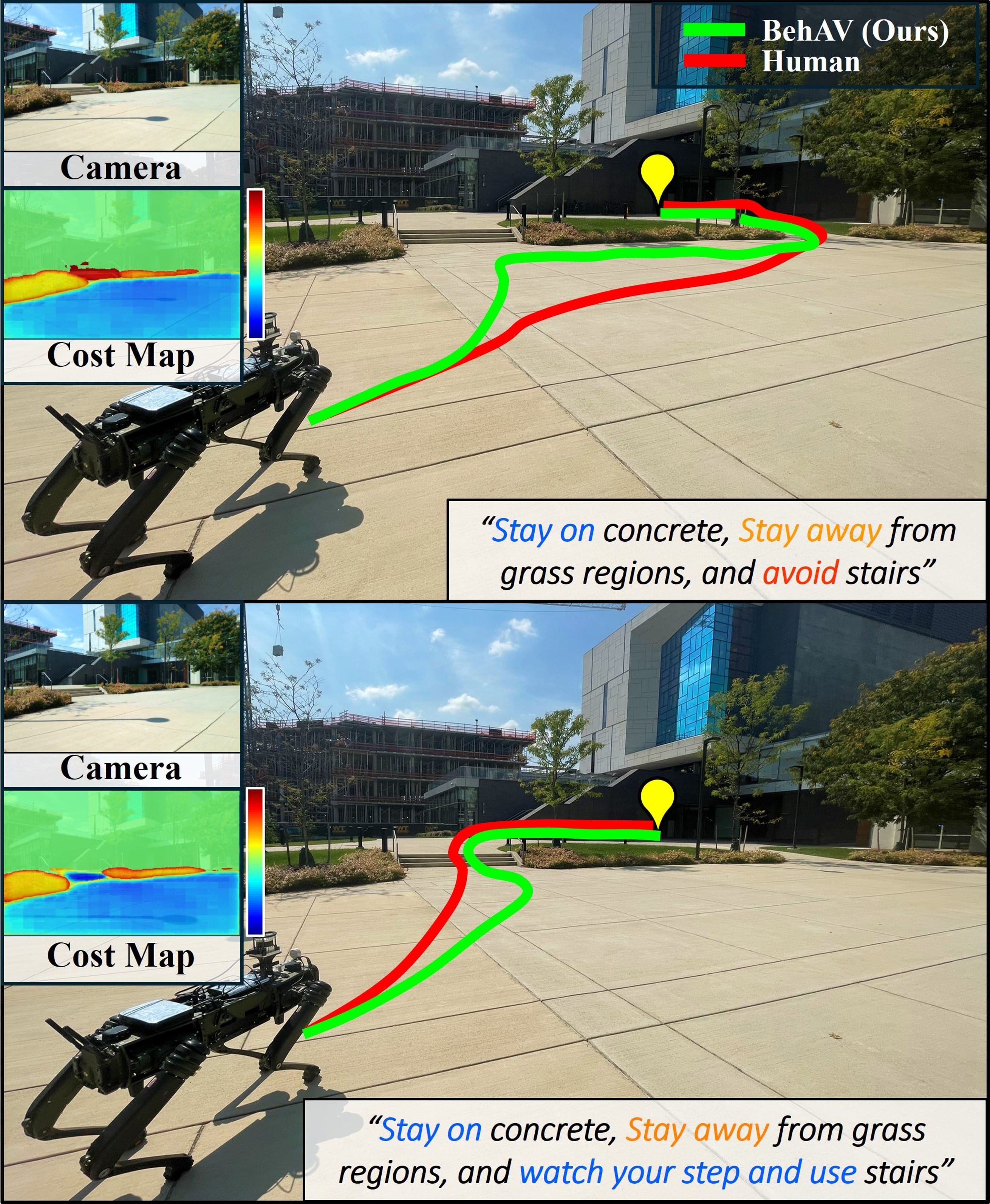}
      \caption {\small{Autonomous robot navigation with \ours{} (ours) for two different behavioral instructions compared to the human preferred path. \ours{} decomposes human instructions into behavioral and navigation instructions. The behavioral instructions are used to construct a real-time behavioral cost map to encode the behavioral rules for planning.   }}
      \label{fig:cover-image}
      \vspace{-10pt}
\end{figure}

Many prior approaches have formulated behavior-aware navigation as a socially compliant navigation task~\cite{liang2021crowd,tang2024encoding}, focusing primarily on standard scene objects such as moving pedestrians, groups of people, vehicles, and lane markers~\cite{9561053}. These approaches typically employ vision and LiDAR-based object detection techniques to identify a predefined set of objects, enabling the imposition of desired navigation behaviors~\cite{GUPTA2024124636}. Various outdoor navigation methods perform terrain traversability estimation, utilizing techniques such as semantic segmentation~\cite{guan2022ganav,10620438}, proactive anomaly detection \cite{weerakoon2022graspe,10417710}, self-supervised learning \cite{sathyamoorthy2022terrapn,10611227}, and representation learning \cite{karnan2023sterling}. These methods aim to enable robots to navigate across diverse terrain types while prioritizing relatively stable terrain regions. Additionally, reinforcement learning \cite{weerakoon2022terp,jiang2024bevnav} and imitation learning \cite{qin2021deep-imitation} have been used to develop navigation policies capable of handling complex terrains and dynamic environments. However, these learning-based methods often face challenges in real-world deployment, particularly in terms of generalization to unseen scenarios and behavioral objectives.


Advancements in Vision Language Models (VLMs)\cite{tang2024chain} and Large Language Models (LLMs)\cite{tang2023saytap} have greatly enhanced zero-shot scene understanding and object detection. Vision-Language Navigation (VLN) has emerged as a promising field, enabling robots to navigate environments using natural language instructions and RGB images~\cite{zhang2024navid, huang2023audio-vLMaps, raj2024rethinking}. VLN utilizes VLMs for tasks like open vocabulary object classification~\cite{10610178}, high-level reasoning~\cite{zhou2024navgpt}, and visual-language grounding~\cite{10650020}, particularly in goal and target detection scenarios. Additionally, vision-language action models~\cite{zitkovich2023rt2} and visual navigation foundation models~\cite{shah2023vint, sridhar2023nomad} have been trained on large-scale datasets to boost performance. Hybrid methods have been developed to combine VLM-based perception with model-based or learning-based planners, where the VLM operates on a remote server~\cite{10342512}. These approaches improve context-aware navigation by providing planners with reference paths~\cite{sathyamoorthy2024convoi}, trajectory selections~\cite{song2024socially}, and waypoint suggestions~\cite{guo2024co} to adhere to behavioral rules. While effective in environments with relatively static behavioral objectives (e.g., crosswalks or terrains), these methods suffer from low inference rates which pose challenges in dynamic scenes (e.g., traffic control signs), where timely responses are crucial for effective navigation.

\textbf{Main contributions:} We present \ours{}, a novel approach that transforms behavioral rules, expressed as natural language instructions, into a real-time behavioral cost map for navigation in outdoor scenes. This behavioral cost map represents both the spatial likelihood of behavioral objects and the costs associated with the preference for actions. We further propose a motion planner to effectively guide robots toward visual targets while adhering to user-defined behavioral instructions. The key contributions of our work include:


\textbf{1. A Novel Behavioral Cost Map Representation} that converts language-based behavioral rules (e.g., \textit{``stop for the stop hand gesture", ``follow the pavement"}) into a cost map using a lightweight VLM that can run on an edge computer. Our method processes RGB images to generate a segmentation map highlighting regions of interest (ROIs) for specified behavioral objects (e.g.,  \textit{``pavement", ``traffic sign"}). The VLM predicts the likelihood of each object being associated with specific ROIs. We use an LLM to estimate the desirability (preference) of the actions linked to these objects, and the scores are mapped to the ROIs, creating a behavioral cost map that integrates both the likelihood of objects and the desirability of actions. 
By grounding these behavioral rules within the cost map, our method enables behavior-aware navigation that can dynamically adapt to changes in the scene in real-time. This flexibility allows the system to perform a variety of behavioral objectives by simply modifying the input language instructions. As a result, our method demonstrates an {15.3}\% increase in behavior-following accuracy compared to existing VLM and vision-based approaches.

\textbf{2. A method for visual goal preprocessing and continuous landmark goal estimation using large VLMs}. We leverage the semantic understanding capabilities of large Vision Language Models (VLMs) to accurately identify landmarks described in language instructions, enabling goal estimation for autonomous robots. In our approach, the camera image is used as an input into a VLM, alongside a language prompt asking for the pixel location of the landmark. The estimated pixel location is then converted into a goal heading direction with respect to the global frame, which is locked as the current goal. This goal direction is continuously re-evaluated using the VLM for any changes. We evaluated the performance of our detection method against a set of ground truth labels and observed an {31.25}\% improvement in F-score and a {37.19}\% reduction in pixel error compared to other VLM-based methods.
%

\textbf{3. A Novel Behavior-aware Planner} to perform landmark goal navigation while adhering to behavioral rules using the generated navigation cost map. The planner introduces a novel objective function to obtain optimal trajectories from an unconstrained model predictive controller. This novel planner produces smooth and contextually appropriate behaviors for actions such as yielding, stopping, and obstacle avoidance. Additionally, we incorporate a behavior-aware gait-switching mechanism that adjusts the robot’s gait during specific behavioral instructions, such as \textit{ \textit{``watch your step."}} The overall planner results in a {40}\% increase in success rate and {22.49}\% closer to human teleoperation in terms of Fréchet distance compared to state-of-the-art methods, highlighting its potential to mimic human-like navigation behavior for a given set of language instructions.


%% file: 2_Related_Work.tex
\section{Related Work} \label{sec:related_work}
In this section, we discuss the existing literature on robot navigation in dynamic scenes, robot navigation using VLMs, and path planning methods.

\begin{figure*}[t]
      \centering
      \includegraphics[width=2\columnwidth]{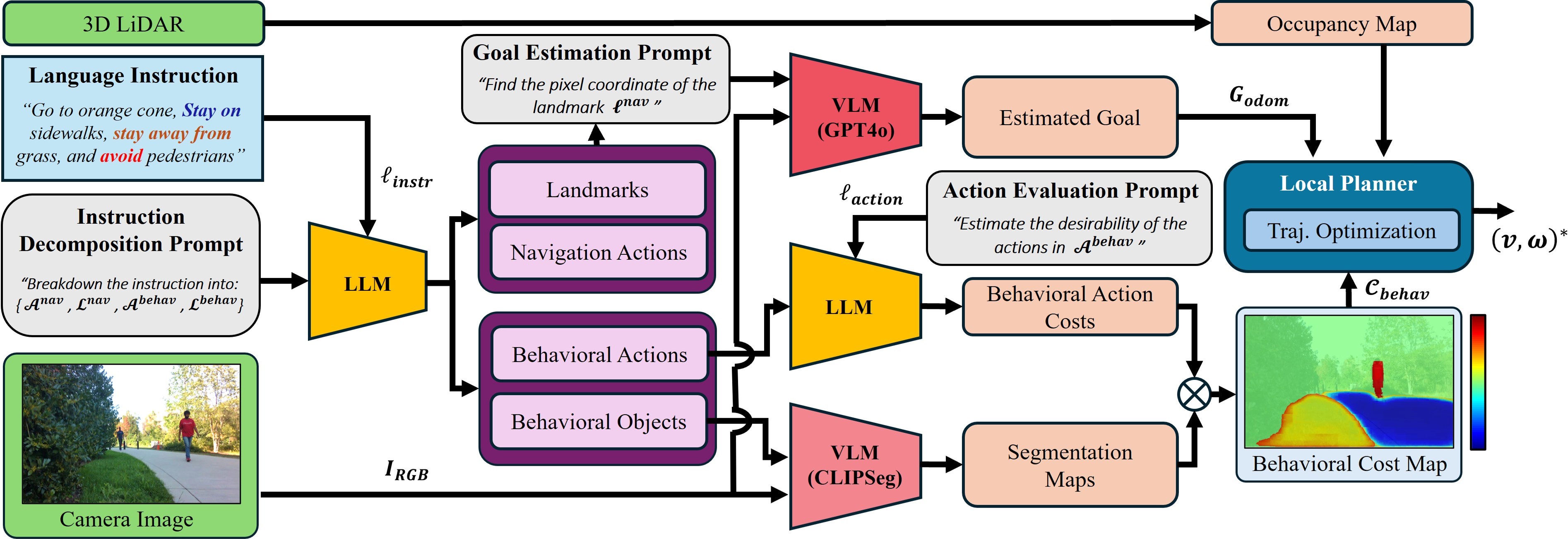}
      \caption {\small{Overall architecture of \ours{}: We decompose human instructions into navigation and behavioral components. Navigation instructions identify landmarks and goals. Behavioral instructions are split into actions ($ \mathcal{A}^{\text{behav}} $) and objects ($ \mathcal{L}^{\text{behav}} $). An LLM evaluates action desirability, assigning probabilities to each action. A lightweight vision-language model (CLIPSeg \cite{luddecke2022clipseg}) generates real-time segmentation maps for behavioral objects. Combining action probabilities with segmentation maps yields a real-time behavioral cost map encoding the instructions. A local planner uses this cost map to navigate toward landmarks while respecting behavioral constraints. }}
      \label{fig:System_arch}
\end{figure*}

\subsection{Robot Navigation in Dynamic Scenes}

Robot navigation in dynamic environments faces two primary challenges: terrain awareness and social interaction \cite{tai2018sociosense}. Terrain-aware navigation enables robots to assess and traverse diverse surfaces in real-time using sensors like IMUs \cite{sathyamoorthy2022terrapn}, proprioception \cite{pronav}, LiDAR \cite{xue2023traversability}, and cameras \cite{frey23fast}. Techniques such as semantic segmentation and self-supervised learning classify terrains as smooth, rough, or non-traversable, optimizing path planning \cite{guan2022ganav,jung2024v,schmid2022self}. Socially aware navigation requires robots to adhere to social norms, predicting human behaviors to maintain safe distances and respect personal space in public areas \cite{liang2021crowd}. Recent methods incorporate attention mechanisms \cite{chen2019crowd} and predictive models \cite{mavrogiannis2018social}, while deep reinforcement learning and imitation learning train robots to navigate around humans without causing discomfort, using simulations and large-scale datasets of human social dynamics \cite{karnan2022socially}. The main challenge is generalizing these learned behaviors to new environments with different social norms or unpredictable pedestrian movements.

\subsection{Robot Navigation Using VLMs}

Recent advancements in large language models (LLMs) and vision-language models (VLMs) have significantly improved scene understanding \cite{jia2024sceneverse}, semantic comprehension \cite{ha2022semabs}, and vision-language grounding for robotic navigation \cite{zhao2024ivlmap}. Early works like CLIP \cite{radford2021clip} and GPT-3 \cite{brown2020gpt3} enabled the extraction and interpretation of landmarks from navigation instructions, where LLMs parsed the textual inputs, and VLMs grounded the landmarks in the robot's environment to guide navigation. These approaches have been extended to develop systems where VLMs plan paths based on spatial references and high-level goals derived from natural language commands. For example, models like ViNT \cite{shah2023vint} and NoMAD \cite{sridhar2023nomad} decompose complex instructions into actionable sub-tasks for task-oriented navigation policies. Recent work on iterative visual prompting using VLMs has further enhanced navigation by continuously updating visual cues as the robot moves \cite{sathyamoorthy2024convoi,song2024socially}. Techniques like open-vocabulary object and goal detection, such as ZSON \cite{majumdar2022zson} and OVExp\cite{wei2024ovexp}, allow for more flexible navigation by identifying unseen objects or goals. Additionally, Navi2Gaze \cite{zhu2024navi2gaze} introduces target gazing strategies, enabling robots to navigate by recognizing and focusing on key targets in dynamic environments. Despite these advancements, the development of adaptive low-level behaviors for dynamic social contexts and terrain changes remains underexplored.





%% file: 3_Background.tex
\section{Background}\label{sec:background}

In this section, we define our notations and provide preliminary details on the unconstrained MPC planner.

\subsection{Notations, and Definitions}

We define three distinct coordinate frames: the odometry frame (odom), which is a fixed global reference frame with its origin and axes set at the robot’s initial position when powered on; the robot frame (robot), which is attached to the robot's center of mass, with the X-axis pointing forward, the Y-axis pointing left, and the Z-axis pointing upward relative to the robot; and the image frame (img), anchored at the top-left corner of the RGB camera image, where the X-axis extends to the right (across columns) and the Y-axis extends downward (across rows). We further assume that the LiDAR frame is identical to the robot frame.

Variables associated with each frame are identified by their subscripts to indicate which frame they belong to; subscript $t$ indicates time stamp (e.g., $x_{\text{rob},t}$ represents a quantity in the robot frame at time $t$). The transformation matrices between these frames are denoted by $\mathcal{T}_{\text{robot}}^{\mathcal{O}}$, $\mathcal{T}_{\mathcal{O}}^{\text{rob}}$, and $\mathcal{T}_{\text{img}}^{\mathcal{O}}$, where each matrix transforms points from the frame indicated by the subscript to the frame indicated by the superscript. Indices $i$ and $j$ are used as notations for relevant points or data elements.



\subsection{Planner Control Policy and Trajectory Parameterization}
\label{sub-sec:traj-parametrization}
In our context, the robot is controlled using linear and angular velocity commands $(v, \omega)$, respectively. Consider a target goal pose \( X_g \), located at a distance \( r \) from the robot.  Let's denote  \( \delta \) as the orientation of the robot relative to the line of sight between the robot and the target. The orientation of the target \( X_g \) relative to this line of sight is represented as $\theta$. 
To guide the robot toward the target pose \( X_g \), \cite{park2012mpepc} proposes a pose-following control law that uses an egocentric goal parameterization \( (r, \theta, \delta) \). The control law computes the angular velocity \( \omega \) needed to steer the robot toward the target and is defined as follows:
\begin{equation}
\omega = \frac{v}{r} \left( k_2 \left( \delta - \text{atan}\left( -k_1 \cdot \theta \right) \right) + \frac{1 + k_1}{1 + k_1^2 \cdot \theta^2} \cdot \sin(\delta) \right).
\label{eq:control-law}
\end{equation}
By choosing a maximum linear velocity $v_{max}$ that controls the curvature of the trajectory, \cite{park2012mpepc, arul2024unconstrained} proposes a 4-dimensional vector $z = (r, \theta, \delta, v_{max})$ that parameterizes a space of smooth trajectories generated by the feedback control policy. 


%% file: 4_Our_Method.tex
\section{ \ours:Behavioral Rule Guided Autonomy Using VLMs} \label{sec:our-method}

In this section, we outline the methodology of our approach. 
\ours{} is structured into four key components: 1. Human Instruction Decomposition; 2. Behavioral Cost Map Generation; 3. Visual Landmark Estimation; 4. Behavior-Aware Planning. Each of these components is discussed in detail below. The overall system architecture is in Fig. \ref{fig:System_arch} and the algorithm is summarized in Algorithm \ref{algo:one}.


\begin{figure}
      \centering
      \includegraphics[width=\columnwidth,height=20cm]{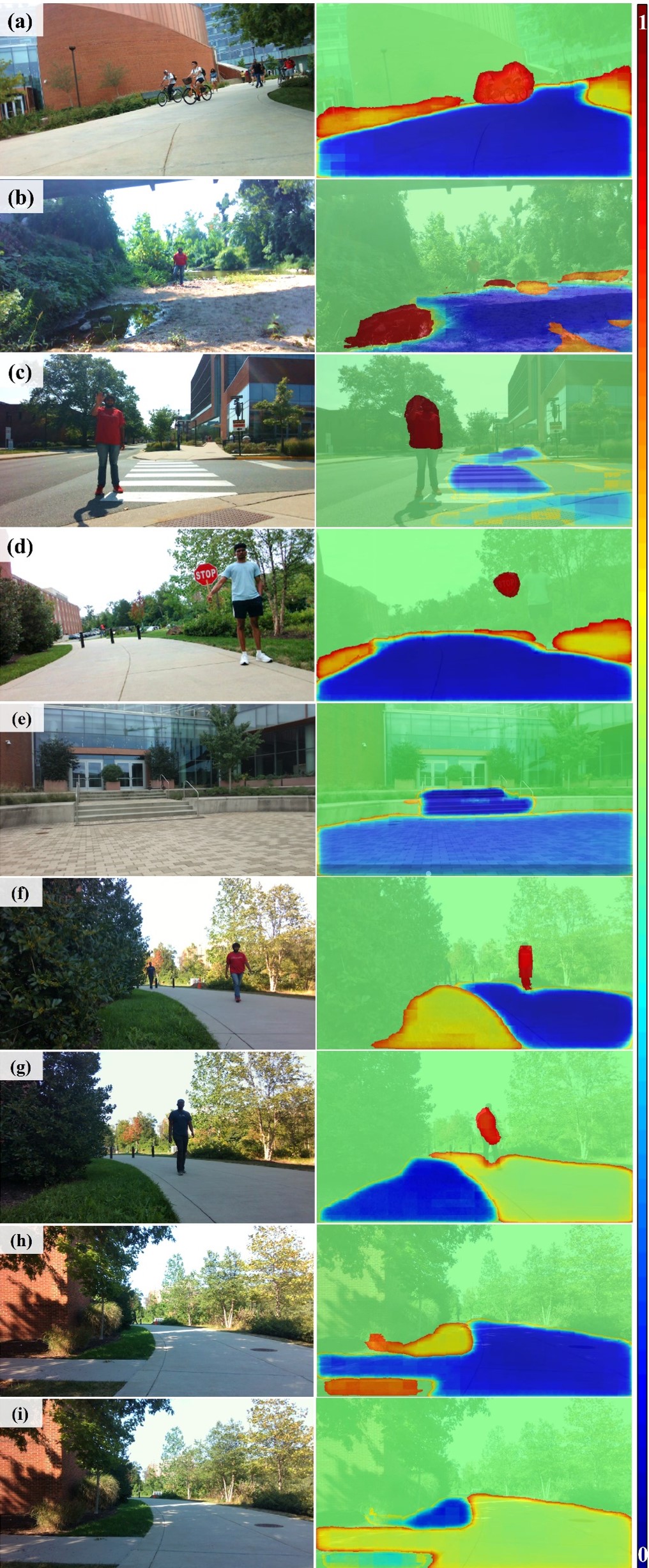}
      \caption {\small{Cost maps generated by \ours{} for diverse instructions: (a)\textit{``Follow the sidewalk, stay away from grass, and avoid cyclists"}; (b)\textit{``Stay on the sand, stay away from grass, and avoid water puddles"}; (c)\textit{``Stay on the sidewalk, follow the crosswalk and stop for stop hand gesture"}; (d)\textit{``Stay on concrete, avoid grass and stop for stop sign"}; (e)\textit{``Stay on tiles, and use caution to follow stairs"}; (f)\textit{``Follow the concrete, stay away from grass, and stop for people wearing red shirts"}; (g)\textit{``Stay on concrete, stay away from grass and yield to people wearing black shirts"}; (h)\textit{``Stay on concrete, stay away from the grass"}; (i)\textit{``Stay on grass, stay away from the concrete"}. The color map is shown on the right side.}}
      \label{fig:behav-cost-map}
\end{figure}


\subsection{Human Instruction Decomposition Using LLMs}
Our method begins by processing high-level human language instructions, $ \ell_{\text{instr}} $, using a predefined prompt, $ \ell_{\text{prompt}} $, as inputs to a large LLM. Specifically,  $ \ell_{\text{instr}} $ is decomposed into four structured sets: Navigation Actions ($\mathcal{A}^{\text{nav}} $),  Navigation Landmarks ($\mathcal{L}^{\text{nav}}$), Behavioral Actions ($ \mathcal{A}^{\text{behav}} $), and Behavioral Targets ($ \mathcal{L}^{\text{behav}} $) using GPT-4. This functional relationship is expressed as:
\begin{equation}
\text{LLM}: \{\ell_{\text{instr}}, \ell_{\text{prompt}}\} \to \{ \mathcal{A}^{\text{nav}}, \mathcal{L}^{\text{nav}}, \mathcal{A}^{\text{behav}}, \mathcal{L}^{\text{behav}} \},
\end{equation}


For example, consider the following human instruction and the corresponding prompt:

$\ell_{\text{instr}}$ = \textit{``Go forward until you see a building with blue glasses, stay on the pavements, stop for stop signs, and stay away from the grass"}

$\ell_{\text{prompt}}$ = \textit{``Given the instruction }$ \ell_{\text{instr}}$ \textit{, list the landmarks (e.g., a building), navigation actions (e.g., go forward), general behavioral actions (e.g., stay on, avoid), and behavioral targets (e.g., pavement) as four separate dictionaries."}

The resulting decompositions are:
$\mathcal{A}^{\text{nav}}$ = \{\textit{``go forward until"}\}; 
$\mathcal{L}^{\text{nav}}$ = \{\textit{``a building with blue glasses"}\};
$\mathcal{A}^{\text{behav}}$ = \{\textit{``stay on", ``stop for", ``stay away from"}\};
$\mathcal{L}^{\text{behav}}$ = \{\textit{``pavements", ``stop sign", ``grass"}\}.


This structured decomposition enables our method to separate high-level landmark goal navigation from local behavioral planning effectively, facilitating behavior-aware robotic navigation. This one-time decomposition process initiates the \ours{} pipeline.

\subsection{Behavioral Action Costs as Conditional Probabilities}
\label{subsec:action_costs}


For each behavioral action in $\mathcal{A}^{\text{behav}}$, we estimate conditional probabilities that describe the likelihood of the action being desirable (a preference) for a given prompt. i.e., for actions such as ``follow" or ``stay on", the probability of being desirable is closer to 1, while for actions like ``stop" the probability of being desirable is closer to 0. These conditional probabilities are obtained by a subsequent query to the large language model (LLM), where the inputs to the LLM are the behavioral action set \( \mathcal{A}^{\text{behav}} \) and a prompt \( \ell_{\text{action}} \). The LLM takes these inputs and maps them to a list of conditional probabilities representing how desirable each action is in the given context. We can represent this mapping as a function:
\begin{equation}
\text{LLM}: \{ \mathcal{A}^{\text{behav}}, \ell_{\text{action}} \} \to \mathcal{P}^{\text{desirable}} = \{ p_1^{\text{desirable}}, \dots, p_n^{\text{desirable}} \},
\end{equation}
\begin{equation}
p_i^{\text{desirable}} = P(\text{desirable} \mid a_i^{\text{behav}} \in  \mathcal{A}^{\text{behav}}, \ell_{\text{action}}),
\end{equation}
where each \( p_i^{\text{desirable}} \) represents the conditional probability that the corresponding behavioral action \( a_i^{\text{behav}} \) is desirable given the action querying prompt \( \ell_{\text{action}} \).


\subsection{Behavioral Cost Map Construction}

We construct a behavioral cost map ($\mathcal{C}_{behav}$) that captures both the probable locations of the behavioral targets and the desirability of the associated behavioral actions. 

\subsubsection{Segmentation Maps for Behavioral Targets}

We first process the input RGB image \( I_{\text{RGB}} \in \mathbb{R}^{H \times W \times 3} \) using the CLIPSeg model \cite{luddecke2022clipseg}, which generates segmentation maps based on behavioral target labels specified in the list \( \mathcal{L}^{\text{behav}} = \{ l_1^{\text{behav}}, l_2^{\text{behav}}, \dots, l_n^{\text{behav}} \} \). (e.g., ``pavement", ``grass", ``road"). The objective is to generate separate segmentation maps for each behavioral target, representing the likelihood that each pixel in the image corresponds to a given behavioral target, thereby providing a spatial distribution of the behavioral objects in $\mathcal{L}^{\text{behav}}$. 

For each behavioral target \( l_i^{\text{behav}} \in \mathcal{L}^{\text{behav}} \), the input image \( I_{\text{RGB}} \) and the corresponding behavioral target \( l_i^{\text{behav}} \) are passed into the CLIPSeg model. The model outputs a single-channel segmentation map \( S_i \in \mathbb{R}^{H \times W} \), where each pixel represents the probability that it belongs to the $i^{th}$ behavioral target:
\begin{equation}
S_i(x, y) = P\left( (x, y) \in l_i^{\text{behav}} \mid I_{\text{RGB}}, \mathcal{L}^{\text{behav}} \right),
\label{eq:clipseg}
\end{equation}
where \( (x, y) \) are the pixel coordinates, and \( S_i(x, y) \) is the probability that pixel \( (x, y) \) belongs to the object specified by \( l_i^{\text{behav}} \).

\subsubsection{Combining Segmentation Maps with Behavioral Action Costs}

For each behavioral target \( l_i^{\text{behav}} \), we assign behavioral action costs to the corresponding pixel locations, while preserving the likelihood distributions obtained from the segmentation maps. To this end, we incorporate the action costs generated as conditional probabilities $\mathcal{P}^{\text{desirable}}$ in section \ref{subsec:action_costs}. Let $\mathcal{C}_{i}$ be the cost map for a given behavioral target $i$. We assign the cost \( c (x,y) = \mathcal{C}_{i}(x,y) \in \mathcal{C}_{i} \) to each pixel $(x,y)$ in $\mathcal{C}_{i}$ as:
\begin{equation}
c (x,y) = p_i^{\text{desirable}}\cdot S_i(x, y), 
\label{eq:cost-combine}
\end{equation}
where, $p_i^{\text{desirable}}$  is the conditional probability that the action $a^{\text{behav}}_i$ (associated with the behavioral target \( l_i^{\text{behav}} \)) is undesirable. \( S_i(x, y) \) represents the likelihood that pixel \( (x, y) \) belongs to \( l_i^{\text{behav}} \). Thus, the cost map $\mathcal{C}_{i}$ reflects both the likelihood of the behavioral target $i$ at every pixel $(x,y)$ and the associated behavioral action cost.

\begin{figure}[t]
      \centering
      \includegraphics[width=1\columnwidth]{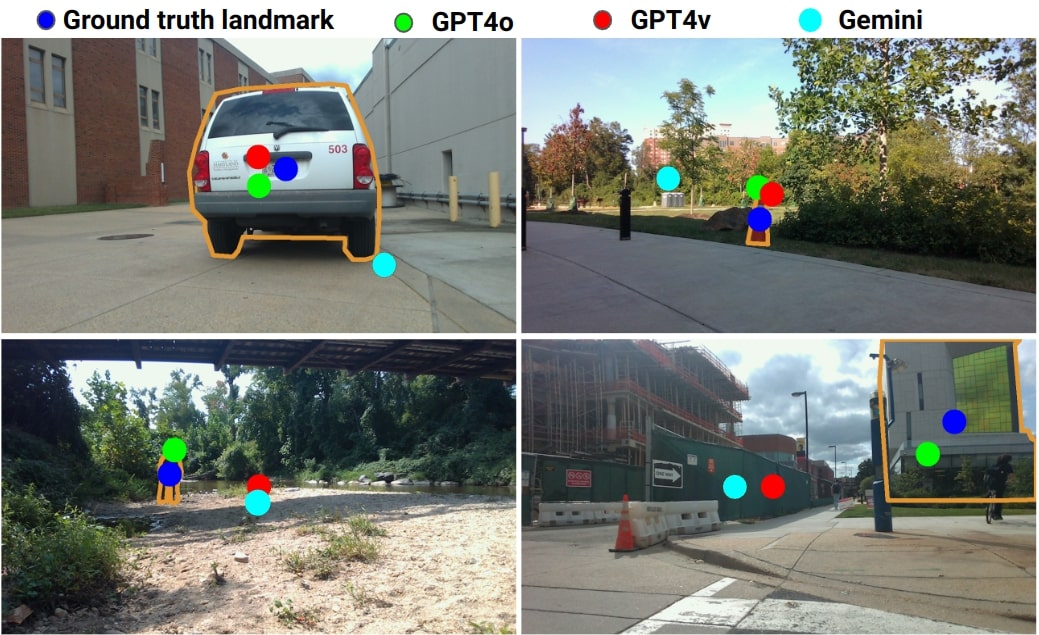}
      \caption {\small{Landmark goal detection using various VLM models compared to the ground truth centroid (blue) across diverse scenes. Predictions from GPT4o (green), GPT4v (red), and Gemini (cyan) are shown, with orange bounding boxes highlighting the landmarks.}}
      
      \label{fig:landmark-detection}
\end{figure}

\subsubsection{Generating the Behavioral Cost Map  ($\mathcal{C}_{behav}$)}

In the presence of multiple behavioral targets influencing the same pixel, we prioritize the target with the highest likelihood and corresponding cost. Hence, the final behavioral cost map $\mathcal{C}_{behav} \in [0,1]^{H \times W}$ is generated by selecting the maximum cost across all behavioral targets for each pixel $(x,y)$ as,
\begin{equation}
\mathcal{C}_{behav}(x, y) = \max_{i=1}^{n} \mathcal{C}_i(x, y),
\label{eq:c-behav}
\end{equation}
where $n$ is the number of behavioral targets in $\mathcal{L}^{\text{behav}}$. Eq. \ref{eq:c-behav} ensures that for each pixel \( (x, y) \), the behavioral target with the highest likelihood and the greatest undesirable action is prioritized. Thus, we maintain the likelihood distributions while ensuring that the most relevant and significant action preference is applied to each pixel.



\begin{figure*}[t]
    \centering
    \includegraphics[width=2\columnwidth]{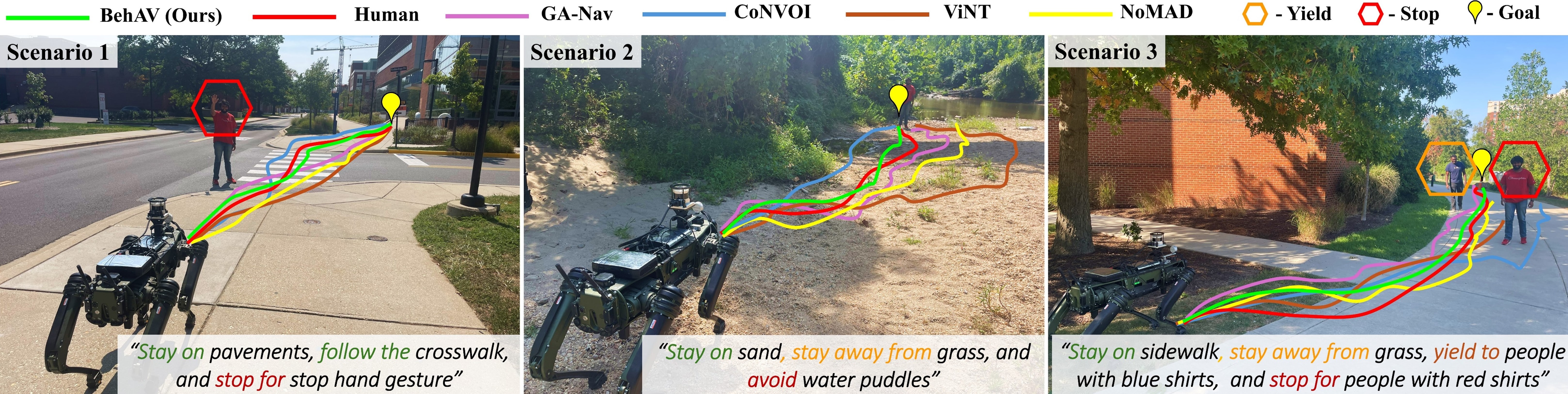}
    \caption{\small{Robot trajectories when navigating in diverse outdoor scenes using various behavioral instructions. \ours{} can demonstrate diverse behaviors by simply changing the input instructions as desired. }}
    \label{fig:comparison_trajs}
\end{figure*}

\subsection{Visual Landmark Estimation}

The robot needs to locate the landmark in $\mathcal{L}^{\text{nav}}$ to follow the navigation actions in $\mathcal{A}^{\text{nav}}$. We utilize vision-language models (VLMs) to identify landmarks from $\mathcal{L}^{\text{nav}}$ and generate navigation goals. Our method takes an RGB image $I_{RGB}$ and a text prompt $\ell_{\text{frontier}}$ to perform visual prompting with GPT-4o VLM, identifying the landmark $l^{\text{nav}}_i$ from $\mathcal{L}^{\text{nav}}$. The VLM outputs the predicted pixel coordinate as:
\begin{equation}
\text{VLM}: \{ I_{\text{RGB}}, \ell_{\text{frontier}}, l^{\text{nav}}_i \} \to G_{img} = [x_{img},y_{img}],
\end{equation}

where $G_{img} \subseteq I_{RGB}$ is the goal location in image coordinates. The corresponding goal in the odometry frame is $G_{odom}^{i} = \mathcal{T}_{img}^{\mathcal{O}}G_{img}^{i}$, where $\mathcal{T}_{img}^{\mathcal{O}} = \mathcal{T}_{robot}^{\mathcal{O}} \mathcal{T}_{img}^{robot}$.

The pipeline handles multiple landmarks by querying the VLM continuously, updating the goal $G_{odom}^{i}$ until the robot reaches each target. Due to internet-based VLM querying, the inference time is $\sim 6$ seconds.

\subsection{Behavior-Aware Planning} \label{sec:mpc-planning}

Our unconstrained model predictive control (MPC) planner leverages the trajectory parameterization \( z = (r, \theta, \delta, v_{\text{max}}) \) detailed in section \ref{sub-sec:traj-parametrization} to optimize a novel objective function that incorporates behavior-aware navigation. Let the robot's look ahead trajectory over a finite time horizon \( T \) be denoted by \( q_z \), parameterized by \( z \). At each planning step, the predictive planner computes \( z^* \), the optimal set of parameters that minimizes the novel cost function \( J \) as follows,
\begin{equation}
z^* = \arg\min_{z} J(q_z),
\end{equation}
where \( J \) is the cost function that evaluates the given \( q_z \) based on three key terms: goal-reaching \( \psi_{\text{goal}}(q_z) \), obstacle avoidance \( \psi_{\text{obs}}(q_z) \), and adherence to behavioral rules \( \psi_{\text{behav}}(q_z) \). The cost function is defined as:
\begin{equation}
J(q_z) = \sum{}  w_{\text{goal}} \cdot \psi_{\text{goal}}(q_z) + w_{\text{obs}} \cdot \psi_{\text{obs}}(q_z) + w_{\text{behav}} \cdot \psi_{\text{behav}}(q_z),
\end{equation}
where $w_i \in \{ \text{goal}, \text{obs}, \text{behav} \}$  are adjustable weights. 

The optimal trajectory parameterization $z^*$ obtained by optimizing the cost function $J(q_z)$ is used to calculate linear and angular velocities $(v,\omega)_t$ for the current time instant $t$ using the control law in Eq. \ref{eq:control-law}.

We calculate goal cost as $\psi_{\text{goal}}(q_z) = \sum_t^T \left( \frac{d_t}{d_{tot}} \right)$, where $d_t$ is the distance to goal from the trajectory location at $t$, and $d_{tot}$ is the line of sight distance from the robot's starting location to the goal location. Then, we define $\psi_{\text{obs}}(q_z)$ as,
\begin{equation}
\psi_{\text{obs}}(q_z) = \sum_{t=1}^{T} \left( \frac{1}{d_{\text{min}, t}} - \frac{1}{d_{\text{safe}}} \right) \cdot \mathbb{I}(d_{\text{min}, t} < d_{\text{safe}}),
\end{equation}
Where, $d_{\text{min}, t}$ is the minimum distance to the obstacle at time $t$, $d_{\text{safe}}$ is the predefined safe distance, $\mathbb{I}(d_{\text{min}, t} < d_{\text{safe}})$ is an indicator function. Hence, $\psi_{\text{obs}}(q_z)$ assigns higher costs to trajectories that come too close to obstacles.

To obtain $\psi_{\text{behav}}(q_z)$, we transform the trajectory $q_z$ to the camera/image frame using perspective projection \cite{park2015perspective-projection} and the coordinate transformation matrix $\mathcal{T}_{\text{robot}}^{img}$. Let $q_z^{img}$ be the trajectory pixel coordinates in the image frame. Then, 
\begin{equation}
    \psi_{\text{behav}}(q_z) = \max \left( \sum_{t=1}^{T} \mathcal{C}_{\text{behav}}(q_{z,t}^{\text{img}}) \cdot e^{-\lambda d_t} \right),
\end{equation}

where, $\mathcal{C}_{\text{behav}}(q_{z,t}^{\text{img}})$ is the behavioral cost at time $t$ of the trajectory $q_z$ the image frame, $d_t$ is the distance of trajectory location at $t$ from the robot, $\lambda$ is a tunable parameter that controls the rate at which the exponential weighting decays. Larger values of $\lambda$ will prioritize closer points more strongly.


\subsubsection{Handling Extremely Undesirable Actions}

When behavioral objects lie beyond the predicted trajectory horizon used to optimize \( J(q_z) \), the planner cannot generate corresponding actions, as these objects are excluded from the optimization. This becomes critical when highly undesirable actions associated with objects beyond the image horizon (e.g., a stop sign) are ignored. To address this, the planner checks for behavioral objects \( l_i^{\text{behav}} \) linked to undesirable actions \( a_i^{\text{behav}} \in \mathcal{A}^{\text{behav}} \). Upon detection, the planner constrains the maximum velocity in the optimizer based on the undesirability of the action. 

Let, $\mathcal{B}_{\text{lower}} = \{ r_{\text{min}}, \theta_{\text{min}}, \delta_{\text{min}}, v_{\text{min}}\}$ and  $\mathcal{B}_{\text{upper}} = \{ r_{\text{max}}, \theta_{\text{max}}, \delta_{\text{max}}, v_{\text{max}}\}$ be the lower and upper bounds of $z$ for the trajectory optimizer, we constrain the $v_{\text{max}}$ in the upper bound as, $v_{\text{max}}' = (1-\text{max}(\mathcal{C}_{behav})) . v_{\text{max}}$ such that the maximum achievable velocity for the robot is significantly lower or zero in the presence of objects associated with extremely undesirable actions (i.e., $\text{max}(\mathcal{C}_{behav}) \geq c_{\text{th}}$, where $c_{\text{th}}$ is a threshold).

This results in reduced velocities according to the behavioral action cost \( p_i^{\text{desirable}} \), ensuring safer navigation even when critical objects are beyond the trajectory horizon.

\subsubsection{Gait Switching for Stability}

Our quadruped robot typically navigates with a default walking gait. To demonstrate the perception capabilities in complex scenarios, we evaluate its performance in scenarios where switching to a more stable gait is necessary to ensure the robot's safety. The planner detects behavioral actions such as \textit{``watch your step''} and \textit{``use caution''}, triggering the robot to lower its center of gravity and adopt cautious stepping when behavioral objects are present, ensuring safe traversal in challenging regions like stairs.


\begin{algorithm}[t]
	\begin{algorithmic}[1]
        \STATE \textbf{Input}: $\ell_{\text{instr}},\ell_{\text{prompt}},\ell_{\text{action}}, \ell_{\text{frontier}}, I_{RGB, t}, P_t^{lidar}, odom$
        \STATE \textbf{Output}: $(v,\omega)_t, gait_{\text{caution}}$
		\STATE \textbf{Initialize} :   $k_1, k_2,T,w_{\text{goal}}, w_{\text{obs}},w_{\text{behav}}, d_{\text{safe}},d_{\text{th}}, \lambda, r_{\text{max}},$ \\ $r_{\text{min}}, \theta_{\text{max}}, \theta_{\text{min}}, \delta_{\text{max}}, \delta_{\text{min}}, v_{\text{max}}, v_{\text{min}}, c_{\text{th}}, gait_{\text{caution}}$
        \STATE $\text{LLM}: \{\ell_{\text{instr}}, \ell_{\text{prompt}}\} \to \{ \mathcal{A}^{\text{nav}}, \mathcal{L}^{\text{nav}}, \mathcal{A}^{\text{behav}}, \mathcal{L}^{\text{behav}} \}$
        \STATE $\text{LLM}: \{ \mathcal{A}^{\text{behav}}, \ell_{\text{action}} \} \to \mathcal{P}^{\text{desirable}}$
        \STATE $\text{VLM}: \{ I_{\text{RGB}}, \ell_{\text{frontier}}, l^{\text{nav}} \} \to G_{img} = [x_{img},y_{img}]$
        \STATE $G_{odom} = \mathcal{T}_{img}^{\mathcal{O}}G_{img}$
        \STATE $d_{\text{goal}} = \|G_{odom} - odom \|_2$

        \STATE $\mathcal{B}_{\text{lower}} = \{ r_{\text{min}}, \theta_{\text{min}}, \delta_{\text{min}}, v_{\text{min}}\}$ \\
        \STATE $\mathcal{B}_{\text{upper}} = \{ r_{\text{max}}, \theta_{\text{max}}, \delta_{\text{max}}, v_{\text{max}}\}$ \\
        
        \WHILE{$d_{\text{goal}} \geq d_{\text{th}}$}
        \STATE $d_{\text{goal}} = \|G_{odom} - odom \|_2$
        \STATE Obtain $S_i \forall l_i^{\text{behav}} \in \mathcal{L}^{\text{behav}}$ from CLIPSeg \cite{luddecke2022clipseg} (Eq. \ref{eq:clipseg})
        \STATE Calculate $\mathcal{C}_{behav}$ using Eq. \ref{eq:cost-combine} and \ref{eq:c-behav}.
        
        \IF{$\text{max}(\mathcal{C}_{behav}) \geq c_{\text{th}}$}
        \STATE $v_{\text{max}}' = (1-\text{max}(\mathcal{C}_{behav})) . v_{\text{max}}$
        \STATE $\mathcal{B}_{\text{upper}} = \{ r_{\text{max}}, \theta_{\text{max}}, \delta_{\text{max}}, v_{\text{max}}'\}$
        \ENDIF
        
        \STATE \textbf{Optimize $J(q_z)$ for $z \in [\mathcal{B}_{\text{lower}},\mathcal{B}_{\text{upper}} ]$}
        \STATE $z^* = \arg\min_{z} J(q_z) $
        \STATE Calculate $(v,\omega)$ for $z^*$ using Eq. \ref{eq:control-law}

        \IF{$ \{\textit{``use caution"} \text{or} \textit{``watch step"} \} \in \mathcal{A}^{\text{behav}}$}
        \STATE $gait_{\text{caution}} = True$
        \ENDIF
        \RETURN $(v,\omega), gait_{\text{caution}}$
        \ENDWHILE
	\end{algorithmic}
	\caption{Outdoor Navigation using BehAV}
	\label{algo:one}
\end{algorithm}

%% file: 5_Results.tex
\section{Results and Analysis}

\subsection{Implementation}
For real-time deployment and inference, we use the Ghost Vision 60 robot from Ghost Robotics equipped with an OS1-32 LiDAR, L515 Realsense Camera, an onboard Intel NUC 11, which includes an Intel i7 CPU and an NVIDIA RTX 2060 GPU. Trajectory optimizer is implemented using the NLopt Python package. We use a combination of global and location optimization to achieve the desired trajectory optimization. Code implementation details with default parameters are available on the \href{http://gamma.umd.edu/behav/}{project webpage}.


\subsection{Comparison Methods and Evaluation Metrics}
We compare our method's navigation performance with CoNVOI \cite{sathyamoorthy2024convoi}, ViNT \cite{shah2023vint}, NoMAD \cite{sridhar2023nomad}, GA-Nav \cite{guan2022ganav}, and DWA \cite{patel2021dwa}. CoNVOI is a context-aware navigation method using VLMs, ViNT is a foundation model for visual navigation, NoMAD is a goal-conditioned diffusion policy, and GA-Nav is an outdoor terrain navigation method that uses semantic segmentation. We perform an ablation study by replacing our planner with the Dynamic Window Approach (DWA) \cite{patel2021dwa}.  For a fair comparison, we provided goal location to the comparison methods that do not take visual targets.


We evaluate navigation perception performance using the following metrics: 



\no \textbf{Success Rate} - The number of times the robot reached its goal while avoiding collisions and \textit{following behavioral rules} over the total number of attempts.

\no \textbf{Avg. Goal Heading Error: } - The average angle error between the goal line of sight and the robot’s heading direction in radians.

\no \textbf{Fréchet Distance \cite{alt1995frechet_distance} w.r.t. Human Teleoperation}: Measures the Fréchet distance \cite{alt1995frechet_distance} (a measure of similarity between two curves) between a human teleoperated path with a comparison method's trajectory. 

\no \textbf{Behavior Following Accuracy (BFA) : } The percentage of the robot’s path length adhered to the behavioral rules out of the totally navigated path length (e.g., the selection of instructed regions/terrains over others and stopping)

We create ground truth labels for a set of landmarks to evaluate our goal-tracking performance with different VLMs using:

\no \textbf{Pixel Error :} Average distance error between the ground truth and predicted landmark pixel location coordinates.

\no \textbf{F-score :} The accuracy of the predicted landmark location is within the ground truth landmark region as an F-score.


\subsection{Testing Scenarios}
We compare our method's navigation performance in real-world outdoor test scenarios that are not included in the training data set. At least 10 trials are conducted in each scenario.

\begin{itemize}
\item \textbf{Scenario 1} - Includes dynamic human gestures, a sidewalk, and crossing the road along a crosswalk.

\item \textbf{Scenario 2} - Multi terrain scenario that includes sand, grass, and water puddles.

\item \textbf{Scenario 3} - Multi-terrain scenario that includes a sidewalk, grass, and vegetation regions along with dynamic pedestrians wearing different colored shirts.

\item \textbf{Scenario 4, 5} - Includes concrete, stairs, and grass regions.

\end{itemize}

\begin{table}
\resizebox{\columnwidth}{!}{%
\begin{tabular}{ c l c c c c c } 
\hline
\textbf{Metrics} & \textbf{Methods} & \multicolumn{1}{p{1cm}}{\centering \textbf{Scn.
1}} & \multicolumn{1}{p{1cm}}{\centering \textbf{Scn.
2}} & \multicolumn{1}{p{1cm}}{\centering \textbf{Scn.
3}} & \multicolumn{1}{p{1cm}}{\centering \textbf{Scn.
4}}  & \multicolumn{1}{p{1cm}}{\centering \textbf{Scn.
5}}

\\ [0.5ex] 
\hline

\multirow{6}{*}{\rotatebox[origin=c]{0}{\makecell{\textbf{Success}\\\textbf{Rate (\%) $\uparrow$ }}}} 
 & GA-Nav \cite{guan2022ganav} & 0 &  70 & 0 & 50 & 20\\
 & ConVOI \cite{sathyamoorthy2024convoi}   & 20 & 50 & 10 & 60 & 50 \\
 & ViNT \cite{shah2023vint}  & 0 & 40 & 0 & 50 & 30\\
 & NOMAD \cite{sridhar2023nomad}& 0 & 30 & 0 & 40  & 20\\
 &  \ours{} with DWA & 80 &  70 &  60 &  60 & 50\\
 &  \ours{} (ours) & \textbf{90} &  \textbf{90} &  \textbf{80} &  \textbf{80} &  \textbf{70}\\
\hline

\multirow{6}{*}{\rotatebox[origin=c]{0}{\makecell{\textbf{Avg. Goal}\\\textbf{Heading $\downarrow$}\\\textbf{Error (rad)}}}} 
 & GA-Nav \cite{guan2022ganav} & 0.652 &  0.598 & \textbf{0.964} & 1.325 &1.345\\
 & ConVOI \cite{sathyamoorthy2024convoi}   & 0.863 & \textbf{0.743} & 0.859& \textbf{1.396} &1.184\\
 & ViNT \cite{shah2023vint}  & 0.746 & 0.654 & 0.687 & 1.257 &1.267\\
 & NOMAD \cite{sridhar2023nomad}& 0.659 & 0.684 & 0.755 & 1.238 &\textbf{1.279}\\
 &  \ours{} with DWA & 0.636 &  0.765 &  0.861 &  1.241 &1.155\\
 &  \ours{} (ours) & \textbf{0.616} &  0.732 &  0.835 &  1.262 &1.137\\
\hline

\multirow{6}{*}{\rotatebox[origin=c]{0}{\makecell{\textbf{Fréchet Dist.}\\\textbf{w.r.t. $\downarrow$ }\\\textbf{Human (m)}}}} 
 & GA-Nav \cite{guan2022ganav}              & 2.188 & 1.015 & 1.584 & 4.251 & 12.357 \\
 & ConVOI \cite{sathyamoorthy2024convoi}    & 1.965 & 1.196 & 1.792 & 6.820 & 8.398 \\
 & ViNT \cite{shah2023vint}                 & 2.235 & 1.258 & 1.458 & 5.783 & 13.774\\
 & NOMAD \cite{sridhar2023nomad}            & 2.291 & 1.234 & 1.522 & 6.207 & 16.247 \\
 &  \ours{} with DWA                        & 1.698 & 1.083 & 1.238 & 3.855 & 6.721\\
 &  \ours{} (ours)                          & \textbf{1.621} & \textbf{0.987} & \textbf{1.195} & \textbf{3.762} & \textbf{5.473}\\
\hline

\multirow{6}{*}{\rotatebox[origin=c]{0}{\makecell{\textbf{Behavior}\\\textbf{Following $\uparrow$ }\\\textbf{Accuracy (\%)}}}} 
 & GA-Nav \cite{guan2022ganav}              & 48.56 &  81.43 & 65.72 & 73.55 & 42.16 \\
 & ConVOI \cite{sathyamoorthy2024convoi}    & 66.32 & 54.11 & 56.02 & 62.84 & 65.32 \\
 & ViNT \cite{shah2023vint}                 & 55.86 & 65.36 & 45.32 & 75.68 & 38.67\\
 & NOMAD \cite{sridhar2023nomad}            & 49.63 & 61.08 & 49.71 & 72.43 & 43.55 \\
 &  \ours{} with DWA                        & 82.90 &  85.72 &  69.65 &  84.67 & 74.63\\
 &  \ours{} (ours)                          & \textbf{86.22} &  \textbf{89.65} &  \textbf{76.14} &  \textbf{88.07} &  \textbf{78.69}\\
\hline

\end{tabular}
}
\caption{\small{Improved navigation performance of \ours{} compared to other methods using various metrics. \ours{} with DWA is an ablation study by replacing our planner with the dynamic window approach \cite{patel2021dwa}.}
}
\label{tab:comparison_table}
\end{table}

\subsection{Analysis and Comparison} \label{sec:analysis}
We present our method results and comparisons qualitatively in Fig. \ref{fig:comparison_trajs}, \ref{fig:cover-image}, and quantitatively in Table \ref{tab:comparison_table}. Ablation studies on landmark goal detection are shown in Table \ref{tab:vlm_comparison}.

\no \textbf{Behavior-aware Navigation Performance:}
We compare all methods to the human-operated trajectory using the Fréchet Distance as a measure of trajectory similarity. \ours{} achieves the lowest Fréchet Distance across all scenarios, indicating its ability to perform human-like, behavior-aware navigation. CoNVOI shows comparable similarity in scenarios 1 and 5 due to its VLM-based scene understanding. GA-Nav performs similarly in multi-terrain scenarios owing to its terrain semantic segmentation capabilities. ViNT and NoMAD consistently reach goals, especially in structured environments; however, their inability to follow the instructed path results in higher Fréchet Distances. 

We further evaluate each method's behavior adherence using the Behavior Following Accuracy (BFA) metric and qualitatively in Fig. \ref{fig:behav-cost-map}. \ours{} achieves approximately 80\% or above adherence across all scenarios, consistently following the instructed behaviors. While ConVOI effectively handles relatively static scenarios like crosswalks, it fails to identify dynamic scene changes such as stop gestures or pedestrians due to its low inference rate. GA-Nav, ViNT, and NoMAD rely on intrinsic navigation behaviors learned during training, enabling them to follow sidewalks and avoid grass; however, these models require fine-tuning or retraining to accommodate new behavioral objectives. Since \ours{} prioritizes following instructed behaviors, it sometimes results in relatively higher goal-heading errors. Furthermore, the inability of other methods to obey behaviors such as stopping and yielding leads to zero success rates in scenarios 1 and 3. In contrast, \ours{} demonstrates consistently higher success rates across all scenarios.

\begin{table}
\resizebox{\columnwidth}{!}{
\begin{tabular}{ l c c c c } 
\hline
\textbf{VLM Model Used} & \multicolumn{1}{p{1.8cm}}{\centering \textbf{Pixel Error $\downarrow$}} & \multicolumn{1}{p{1.2cm}}{\centering \textbf{F-Score} $\uparrow$} & \multicolumn{1}{p{2cm}}{\centering \textbf{Infer. Rate(sec)}}\\ [0.5ex] 
\hline

Gemini  & 274.41 & 0.44 & 10.65   \\
GPT4v  & 245.49 &  0.433 & \textbf{3.37}   \\
GPT4o & \textbf{178.94} & \textbf{0.64} & 3.59 \\
\hline
    
\end{tabular}
}
\caption{\small{Landmark detection performance of various large VLMs using the prompting strategy proposed in \ours{}. The table compares the pixel error, F-score, and inference rate for each model. A total of 250 images across diverse scenes were used, featuring 7 different landmarks to compare against the ground truth labels.}}
\label{tab:vlm_comparison}
\vspace{-20pt}
\end{table}

\no \textbf{Landmark Detection Performance:}
We compare our landmark detection strategy's performance in \ref{fig:landmark-detection} and Table \ref{tab:vlm_comparison}. We observe that GPT4o demonstrates superior landmark estimation accuracy and a higher inference rate compared to other VLMs. We evaluate detection accuracy across varying landmark distances and perspectives, observing that GPT4o provides accurate visual reasoning for given language descriptions of landmarks.

\no \textbf{Ablation on Behavior-aware Planner: } We perform an ablation study by replacing our proposed planner with the DWA \cite{patel2021dwa} planner to highlight the effect of unconstrained MPC stratergy. We observe that \ours{}'s planner results in smoother trajectories compared to DWA leading to lower goal heading error.

\no \textbf{Inference Rate:} Our overall system runs $\sim 9.5 Hz$ on  Intel NUC 11 which makes it effective to use in dynamic scene handling during robot navigation. 

%% file: 6_Conclusions.tex
\section{Conclusions, Limitations and Future Work}
We introduced \ours{}, a novel approach for autonomous robot navigation in dynamic environments, guided by human instructions and enhanced by Vision-Language Models (VLMs). Our system uses a Large Language Model (LLM) to interpret commands into navigation and behavioral guidelines, integrating language understanding with visual perception. Leveraging VLMs for zero-shot scene understanding, we estimate landmark locations from RGB images. Our behavior-aware planner prioritizes reaching landmarks while adhering to behavioral guidelines. Experimental validation on a quadruped robot in diverse real-world scenarios demonstrates significant improvements over state-of-the-art methods. Limitations include susceptibility to lighting conditions affecting scene understanding and LLM/VLM prediction errors due to hallucinations. Future work aims to enhance perception robustness using advanced multi-modal language models.